\documentclass[journal]{IEEEtran}

\makeatletter
\let\MYcaption\@makecaption
\makeatother

\usepackage[font=footnotesize]{subcaption}

\makeatletter
\let\@makecaption\MYcaption
\makeatother

\usepackage{booktabs}  
\usepackage{threeparttable}
\usepackage{multirow}
\usepackage{tabularx}

\usepackage{amsmath}

\usepackage{hyperref}

\usepackage{xcolor}

\usepackage{graphicx}
\graphicspath{ {figures/} }

\begin{document}
%
\title{Edge Prior Augmented Networks for \\ Motion Deblurring on Naturally Blurry Images}
%
%
%

\author{Yuedong~Chen{$^\top$},
        Junjia~Huang{$^\top$},
        Jianfeng~Wang,
        and~Xiaohua~Xie{$^\ast$}
\thanks{{$^\top$}~The first two authors contributed equally to this work.}%
\thanks{{$^\ast$}~Xiaohua Xie is the corresponding author.}%
\thanks{Y. Chen (yuedong.chen@monash.edu) is with Moansh University, Australia. J. Wang (jianfeng.wang@cs.ox.ac.uk) is with University of Oxford, UK. J. Huang (huangjj87@mail2.sysu.edu.cn) and X. Xie (xiexiaoh6@mail.sysu.edu.cn) are with the School of Computer Science and Engineering, Sun Yat-sen University, Guangzhou 510006, China.}
}

%
%

%

\maketitle

\begin{abstract}
Motion deblurring has witnessed rapid development in recent years, and most of the recent methods address it by using deep learning techniques, with the help of different kinds of prior knowledge. Concerning that deblurring is essentially expected to improve the image sharpness, edge information can serve as an important prior. However, the edge has not yet been seriously taken into consideration in previous methods when designing deep models. 
To this end, we present a novel framework that incorporates edge prior knowledge into deep models, termed \textbf{Edge Prior Augmented Networks (EPAN)}. EPAN has a content-based main branch and an edge-based auxiliary branch, which are constructed as a Content Deblurring Net (CDN) and an Edge Enhancement Net (EEN), respectively. EEN is designed to augment CDN in the deblurring process via an attentive fusion mechanism, where edge features are mapped as spatial masks to guide content features in a feature-based hierarchical manner. An edge-guided loss function is proposed to further regulate the optimization of EPAN by enforcing the focus on edge areas.
Besides, we design a dual-camera-based image capturing setting to build a new dataset, Real Object Motion Blur (ROMB), with paired sharp and naturally blurry images of fast-moving cars, so as to better train motion deblurring models and benchmark the capability of motion deblurring algorithms in practice. Extensive experiments on the proposed ROMB and other existing datasets demonstrate that EPAN outperforms state-of-the-art approaches qualitatively and quantitatively. 

\end{abstract}


%
\IEEEpeerreviewmaketitle

\section{Introduction}

\IEEEPARstart{M}{otion} blur is caused by the relative movement between the camera and objects during the exposure time, either due to the long exposure time of the camera~\cite{nayar2004motion} or the fast-moving velocity of objects~\cite{raskar2006coded}. It is mainly shown as a striking trail along the trajectory of the object~\cite{navarro2011motion}. 

Removing motion blur flaws from a degraded image is inherently ill-posed and challenging, yet it attracts great attention within the computer vision and graphics community for decades. Conventional deblurring algorithms mainly focus on regularizing the solution space by introducing different image priors~\cite{navarro2011motion,chan1998total,cho2009fast,goldstein2012blur,xu2010two}. With the rapid development of convolution neural networks (CNNs), researchers have demonstrated the effectiveness of CNNs in the motion deblurring task. Some earlier related works mainly leverage CNNs to assist conventional kernel-based deblurring algorithms~\cite{xu2014deep,sun2015learning,gong2017motion}. More recently, deep learning-based approaches tend to directly deblur the input image via training with sharp-blurry image pairs, which generally take into consideration different kinds of prior knowledge, such as multi-scale~\cite{nah2017deep,tao2018scale,gao2019dynamic}, blur features~\cite{lu2019unsupervised}, local patch~\cite{zhang2019deep,suin2020spatially}, optical flow~\cite{yuan2020efficient}, and image contents~\cite{shen2019human,kupyn2018deblurgan}.


Although many kinds of priors prove to be helpful, our intriguing finding is that edge information is much more closely related to deblurring. In general, deblurring aims at improving the image sharpness, which is mainly affected by two factors, namely \textit{resolution} and \textit{acutance}~\cite{maitre2017photon}. While the resolution has been mainly addressed by the image super-resolution task, 
acutance, related to the edge contrast, is worth noticing for image deblurring. 
Such edge information is frequently leveraged as an important prior in traditional approaches, where high-frequency contents are detected to help optimize the motion kernels~\cite{xu2010two,cho2009fast,shan2008high}.
However, regardless of its promising potential in boosting the deblurring performance, edge information has not yet been well adopted by any CNNs based deblurring methods. 
To fill the gaps, we propose a novel deep learning-based motion deblur framework, termed Edge Prior Augmented Networks (EPAN), which explicitly models and embeds the edge information in the feature space.
As depicted in Fig.~\ref{fig:model_archi}, the EPAN consists of two branches with similar structures, namely Content Deblurring Net (CDN) and Edge Enhancement Net (EEN). 

Given a blurry input image $I_B$, an existing edge detection algorithm, \textit{e.g.}, canny edge detector~\cite{canny1986computational}, is used to obtain the corresponding blurry edge map $M_B$. While $M_B$ goes into the EEN to get its edge enhanced, the blurry input image $I_B$ is fed into the CDN. 
Concerning that the intermediate features from the EEN decoder are edge-sensitive, they are further leveraged to augment the decoding process of the CDN.
This is achieved by an attentive fusion mechanism, where the edge features are mapped as spatial attention masks and applied to weigh the content features in a feature-based hierarchical manner. Due to the importance of edge information, we also propose an edge-guided loss function to guide the training of both CDN and EEN.

We notice that a dataset with both naturally blurry and sharp images is of vital importance for training a reliable motion deblurring model since manually generated blurry images are different from real ones~\cite{zhang2020deblurring}. Researchers have spent great effort moving toward real image datasets. Most existing works resort to simulating blurry images by averaging consecutive frames of sharp images~\cite{nah2017deep,shen2019human}. More recently, a dataset with both real blurry and sharp images for static or low-speed moving objects is introduced~\cite{rim2020real}.
Following this trend, we move one step forward by introducing a dataset with real sharp and blurry image pairs, focusing on high-speed moving objects. The data are collected using a dual-camera setting, where one camera is set with a shorter exposure time to capture sharp images, while the other is set with a longer exposure time to capture corresponding blurry images. For simplicity and practicality, our dataset contains only fast-moving cars and it is named Real Object Motion Blur (ROMB).

Our contributions are mainly three folds, 
\begin{itemize}
  \item We pinpoint the importance of edge information in motion deblurring and propose a novel deblurring model, EPAN, by explicitly modeling edge information as prior knowledge. To our knowledge, this is the first work to incorporate edge information into CNNs for motion deblurring. 
  \item We propose a new dataset with real sharp-blurry image pairs with fast-moving cars using a dual-camera setting.
  \item Extensive ablation studies verify the effectiveness of different components of EPAN, and experiments on the proposed ROMB and other benchmark datasets demonstrate that EPAN outperforms state-of-out-art deblur models quantitatively and qualitatively.
\end{itemize}

The remaining parts of the paper are structured as follows. Related image deblurring datasets and algorithms will be introduced in Section~\ref{related}. In Section~\ref{model}, we will elaborate the model architecture and training loss functions of the proposed deblurring approach. Details of the proposed dataset will be given in Section~\ref{dataset}. In Section~\ref{experiments}, we will showcase the results and analysis of the ablation study on the proposed model and compare the proposed model with other state-of-the-art deblurring models on several benchmarks. Finally, we will conclude our work and point out the future research direction in Section~\ref{conclusion}.

\section{Related Work} \label{related}
In this section, we mainly review two topics that are closely related to our work, including motion deblurring approaches and some existing benchmark datasets designed for motion deblurring. 

\subsection{Motion Deblurring Algorithms}
Conventional motion deblurring approaches mainly started by constructing energy functions, which would be optimized to transform the blurry images into latent sharp ones~\cite{rajagopalan2014motion}. Specifically, edge information was treated as the important prior knowledge. For example,
Jia \textit{et al.}~\cite{jia2007single} proposed an edge selection algorithm, where motion kernels were estimated by using the calculated edge transparency. Xu \textit{et al.}~\cite{xu2010two} improved the algorithm by introducing mask computation to adaptively select only those effective edges.
While edge information proved to be useful, these conventional approaches were time-consuming due to the requirement of iterative optimization. 

In the past decade, deep learning models have gradually dominated many computer vision tasks~\cite{chen2019facial, chen2020label}, similarly, they have also seen 
a great breakthrough in motion deblurring problem. Sun \textit{et al.}~\cite{sun2015learning} mixed a CNN model with Markov random field to learn and refine patch level motion kernels, which would be used to reconstruct the latent sharp images.
Follow this setting, Gong \textit{et al.}~\cite{gong2017motion} exploited Fully Convolutional Networks (FCN)~\cite{long2015fully} to directly learn a kernel map. However, these approaches were subjected to the precision of kernel estimation, because even small kernel errors could give rise to significant artifacts in the generated latent sharp image~\cite{shan2008high}.

More recently, a trending solution was to design kernel free end-to-end model, mainly with help from different kinds of prior knowledge. 
Nah \textit{et al.}~\cite{nah2017deep} proposed a multi-scale CNN model to mimics conventional coarse-to-fine approaches. 
Such a multi-scale structure was later extended to be a multi-patch hierarchical networks, where the deeper level sub-networks divided the input images into more patches~\cite{zhang2019deep}. And it was further extended by adding attention modules~\cite{suin2020spatially}.
Kupyn \textit{et al.}~\cite{kupyn2018deblurgan, kupyn2019deblurgan} explored using GAN to tackle motion deblurring tasks. 
Zhang \textit{et al.}~\cite{zhang2020deblurring} combined a blurring GAN with a deblurring GAN, assuming that learning to blur could benefit the deblurring process. 
Recently, Gao \textit{et al.}~\cite{gao2019dynamic} designed a multi-scale deblur framework with independent sub-networks, due to the observation that edge information was different across different scales. These approaches overlook the relationship between the motion blur and edge information, and the research on motion deblurring is stuck at a bottleneck.  

\begin{figure*}
    \centering
    \includegraphics[width=0.98\textwidth]{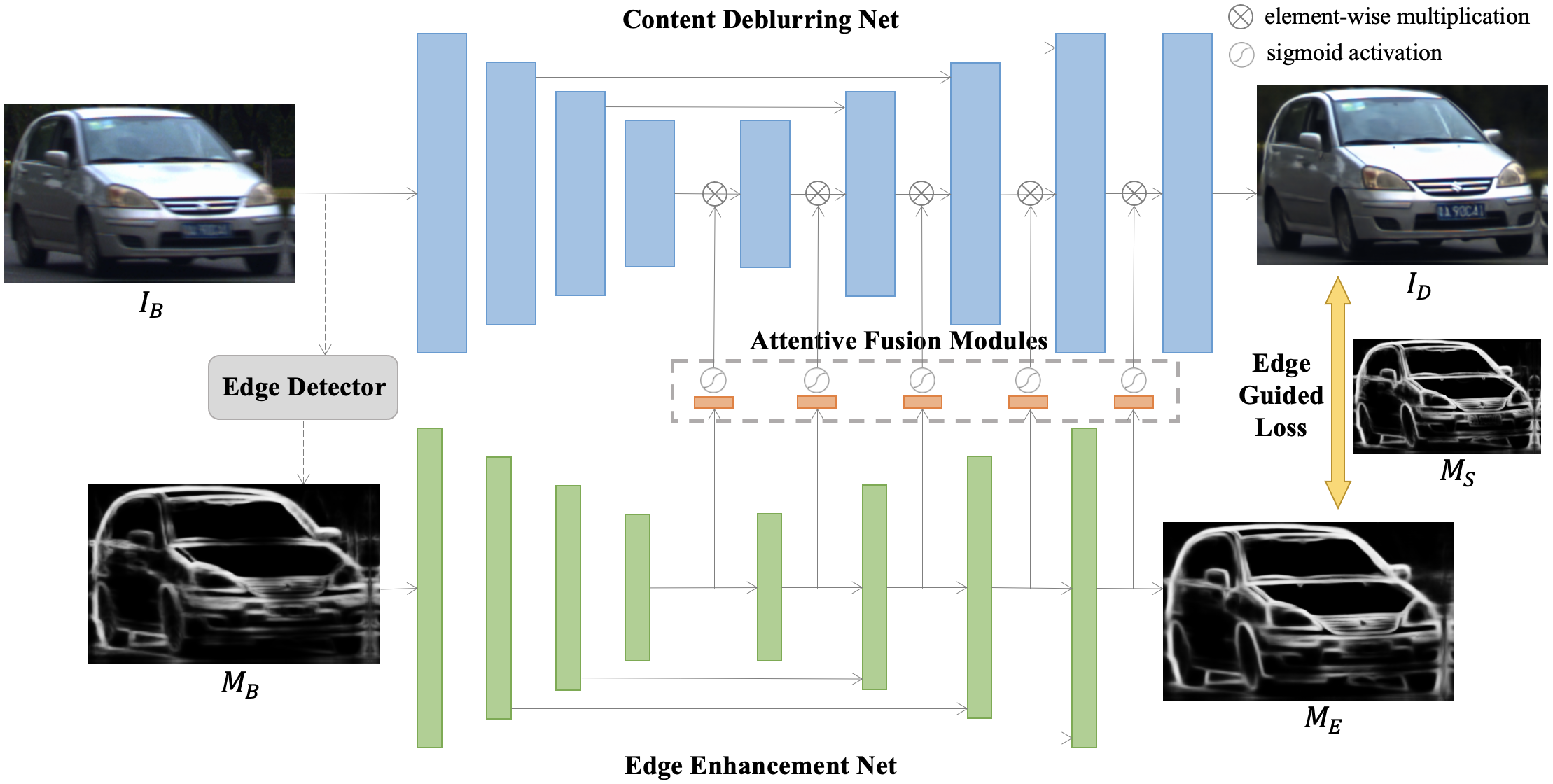}
    \caption{The architecture of Edge Prior Augmented Networks (EPAN), where $I_B$, $I_D$, $M_B$, $M_E$ and $M_S$ refer to blurry image, deblurred image, blurry edge map, enhanced edge map and sharp edge map, respectively. EPAN consists of a Content Deblurring Net (CDN) and an Edge Enhancement Net (EEN). Specifically, EEN provides CDN with edge prior knowledge to augment the deblurring process via attentive fusion modules. Besides, an edge guided loss is also proposed to further enhance the edge information. 
    }
    \label{fig:model_archi}
\end{figure*}

In this work, we intriguingly find out that edge information can be leveraged as the useful prior knowledge for deep learning based model. By proposing an edge augmented deblurring framework, we verify that our design is more closely related to motion deblurring and can handle real blurry images more effectively.

\subsection{Datasets for Motion Deblurring}

Datasets for motion deblurring mainly contain blurry images and the corresponding sharp images, while motion kernels may also be provided in some cases. 

Due to the limitation of devices and computational capacity, blurry images in most early day datasets were artificially synthesized, achieved by convoluting the original sharp images with single motion kernel~\cite{levin2009understanding}, mixed linear kernels~\cite{sun2015learning}, varied types of kernels~\cite{lai2016comparative}, 
kernels generated using captured trajectory~\cite{kupyn2018deblurgan}, \textit{etc}.
However, unlike naturally blurry images, synthetic blurry images did not contain scene depth variation, sensor saturation and unknown camera noise, \textit{etc}.~\cite{lai2016comparative}. Thus, algorithms that performed very well on synthetic datasets often failed to deblur images captured on real scenes.

Because of the aforementioned drawbacks of synthetic data, capturing naturally blurry images raised great attention. Early attempts~\cite{kohler2012recording, lai2016comparative} collected few samples by recording the detailed 6D camera trajectories, followed by playing them back on a robotic platform.
Nah \textit{et al.}~\cite{nah2017deep} chose to approximate integration by summation on time domain, where they obtained blurry images by averaging a short clip of sharp images. Following the same settings, Shen \textit{et al.}~\cite{shen2019human} curated a new dataset with a special focus on human faces.
More recently, by using a beam splitter, Rim \textit{et al.}~\cite{rim2020real} built up an image acquisition system to capture both real blurry and sharp images at the same time, and they collected their data by moving the camera, resulting in camera motion blur for static objects.

Inspired by the above progress, we propose a dual camera based scheme that is easy-to-set-up and low cost. Based on this scheme,
we construct a new dataset with fast-moving objects for motion deblurring, where sharp and blurry images are both captured by cameras directly and have an one-to-one correspondence.

\section{Proposed Deblurring Approach} \label{model}

Motion deblurring generally aims to solve $P(I_D | I_B)$, where $I_B$ and $I_D$ indicate the input blurry image and the reconstructed deblurred image, respectively. Considering that edge information is closely related to the image sharpness and can be leveraged as prior knowledge as aforementioned,
our proposed Edge Prior Augmented Networks (EPAN) choose to solve $P(I_D | I_B, M)$, where $M$ refers to the edge map information. In this section, we will elaborate how edge information is embedded into the architecture of EPAN (Section~\ref{net_archi}), how edge information can help to enhance the optimization of EPAN (Section~\ref{obj_fun}), and how EPAN does inference in the testing phase (Section~\ref{model_infer}).

\subsection{Networks Architecture} \label{net_archi}

The process of human beings to handle the image motion deblurring can be roughly divided into two steps.
Firstly, the semantic information of the image needs to be comprehended, and secondly, a deblurred image is generated by reconstructing the semantic information with sharp edges. 
We design our deblurring framework EPAN by mimicking the above human activity.
As illustrated in Fig.~\ref{fig:model_archi}, EPAN consists of two branches, including \textbf{Content Deblurring Net (CDN)} and \textbf{Edge Enhancement Net (EEN)}. 

CDN is the main branch, which recovers a deblurred image when given a blurry image. CDN is supposed to extract high dimension semantic features, followed by reconstructing them back into the low dimension image space, hence the encoder-decoder structure is a good choice to achieve this goal. Since 
high dimension semantic features contain very little structural information, simply decoding them may fail to yield high-quality outcomes or even cause model collapse. As a general approach~\cite{ronneberger2015u}, we add skip connections between the encoder and the decoder, so as to provide more structural information into the decoding process. However, because the original input is a blurry image, its structural information may be distorted. Is it possible to augment the content decoder with further enhanced information? We, therefore, introduce the EEN, together with an effective fusion mechanism. 

EEN is the auxiliary branch, which aims to enhance the quality of a corresponding blurry edge map, so as to provide augmented decoding guidance for CDN. The blurry edge map can be obtained directly from the blurry input image by using any off-the-shelf edge detector, \textit{e.g.}, canny edge detector~\cite{canny1986computational}, HED~\cite{xie2015holistically}, we will detail our implementation in the experiments section (Section~\ref{exp:setting}). Considering the similar objectives between CDN and EEN, we minor the model architecture of CDN to construct EEN. Besides, the edge map contains less information than the original image, enhancing the edge map should be an easier task, thus we choose to shrink the channel number of each building block of EEN to 1/4 of the corresponding one of CDN.

To make sure that edge information can be effectively employed in the deblurring process, we adopt an \textbf{Attentive Fusion Mechanism} to fuse the edge feature with the content feature. Note that fusion is only applied to the decoders, as we do not intend to alter the content semantic information extracted by the CDN encoder.
The building block of the mechanism is the attentive fusion module, which is given as
\begin{equation}
f^{(i)}_{\text{AttFu}}(x) = \sigma (g(x^{(i)}_E)) \otimes x^{(i)}_C, 
\end{equation}
where $x^{(i)}_E$ is the $i$-th layer feature extracted by the decoder of EEN, while $x^{(i)}_C$ is the corresponding $i$-th layer feature extracted by CDN; $g(\cdot)$ is a trainable mapping network, $\sigma$ is the sigmoid function, and $\otimes$ refers to element-wise multiplication. In our experiment, we model $g(\cdot)$ with a single convolution layer, whose output channel size is set to 1. Intuitively, the attentive fusion module aims to convert the input edge feature into a spatial attention mask, which will further attend to the decoder of CDN to enhance the edge information in the content deblurring process.
Inspired by the pyramid representation and the differences of edge area among different scales~\cite{gao2019dynamic}, we choose to fuse the edge feature $x^{(i)}_E$ with content feature $x^{(i)}_C$ on every single convolution layer of the decoders (see Fig.~\ref{fig:model_archi}).  

\subsection{Objective Functions} \label{obj_fun}

\noindent\textbf{Objective Functions for CDN.} 
In order to improve the sharpness of an image, the traditional image sharpening algorithm chooses to enhance the edge contrast, by designing a filter/mask to place more focuses on the edge boundaries.
Inspired by this, we propose an \textbf{Edge Guided Loss} function. Specifically, the edge map is leveraged as a weight map to control the contribution of each pixel in the network output, which can be written as,
\begin{equation} \label{eq:edge}
l_{\text{edge}}(I_D, I_S, M_S) = -\frac{1}{n}\sum_{i}^{n}M_S^i(I_D^i - I_S^i)^2,
\end{equation}
where $I_S$ and $I_D$ refer to the ground truth sharp image and the deblurred image generated by CDN given a blurry image $I_B$, respectively. $M_S$ denotes the sharp edge map extracted from $I_S$, and $n$ is the number of pixels in each image, while $i$ is the pixel index. It is reasonable to use the sharp edge map $M_S$ rather than the generated enhanced edge map $M_E$ as the mask because $M_S$ is accessible in training and it provides much more accurate and stable information than $M_E$.

Since the value becomes 0 in the non-edge area in $M_S$, using only Eq.~\eqref{eq:edge} would omit all contributions from the non-edge smooth region. The fact is that those smooth region still contains lots of important information, such as color and texture, hence an additional constraint is needed. Similar to the previous motion deblurring approaches~\cite{tao2018scale, gao2019dynamic}, we add the mean square error constraint, which is,
\begin{equation} \label{eq:mse}
l_{\text{mse}}(I_D, I_S) = -\frac{1}{n}\sum_{i}^{n}(I_D^i - I_S^i)^2.
\end{equation}

In general, the loss function for CDN is a weighted combination of both $l_{\text{edge}}$ and $l_{\text{mse}}$, which is,
\begin{equation} \label{eq:cdn}
\begin{split}
    l_{\text{CDN}}(I_D, I_S, M_S) & = \lambda_{\text{C}} \cdot l_{\text{edge}} + l_{\text{mse}} \\
& = -\frac{1}{n}\sum_{i}^{n}(\lambda_{\text{C}} \cdot M_S^i + 1)(I_D^i - I_S^i)^2,
\end{split}
\end{equation}
where $\lambda_{\text{C}}$ is a hyper-parameter, and all other notations refer to the same ones as in Eq.~\eqref{eq:edge}.

\noindent\textbf{Objective Functions for EEN.} Similar to the above CDN branch, the objective functions for EEN is given as,
\begin{equation} \label{eq:een}
l_{\text{EEN}}(M_E, M_S)   
 = -\frac{1}{n}\sum_{i}^{n}(\lambda_{\text{E}} \cdot M_S^i + 1)(M_E^i - M_S^i)^2,
\end{equation}
where $M_E$ denotes the enhanced edge map generated by the EEN branch given a blurry edge map $M_B$, and $\lambda_{\text{E}}$ is another hyper-parameter similar to $\lambda_{\text{C}}$ in Eq.~\eqref{eq:cdn}.

\noindent\textbf{Total Objective Functions.} In general, the objective functions used for optimizing EPAN is given as,
\begin{equation}
\begin{split}
l_{\text{EPAN}}(I_D, I_S, M_E, M_S)   
  = & l_{\text{CDN}}(I_D, I_S, M_S) \\
   & + l_{\text{EEN}}(M_E, M_S),
\end{split}
\end{equation}
where $l_{\text{CDN}}$ and $l_{\text{EEN}}$ are detailed in Eq.~\eqref{eq:cdn} and \eqref{eq:een}, respectively. During training phase, CDN and EEN are optimized simultaneously through minimising $l_{\text{EPAN}}$. 

\subsection{Model Inference} \label{model_infer}
During the testing phase, the given blurry image will go through the same two branches structure, leveraging the same Edge Detector as used in the training phase. However, only the output of CDN, \textit{i.e.}, deblurred image, will be kept for further evaluation. The edge maps from the auxiliary EEN branch are mainly designed to augment the image deblurring process from the feature domain, and they will \textit{not} be used to refine the generated deblurred images via any post-processing method. Therefore, these edge maps can be simply discarded in testing.

\section{Proposed Dataset} \label{dataset}
Recent proposed deblurring datasets~\cite{nah2017deep, shen2019human} mainly synthesized the blurry images by averaging several consecutive frames of sharp images, but there are domain gaps between synthetic blurry images and naturally blurry ones. To mitigate such domain gaps, an existing work~\cite{rim2020real} designed an image acquisition system with a beam splitter to capture both sharp and blurry images at the same time, but the system was somewhat complex and the constructed dataset mainly included static objects. Therefore, to better benchmark motion deblurring in real scenes, we choose to build an easy-to-set-up dual-camera capturing system and curate a dataset with a special focus on fast-moving cars. In this section, we will detail our proposed dataset, including how the images are collected (Section~\ref{data:collect}), how we clean the data (Section~\ref{data:clean}), and how the dataset is organized (Section~\ref{data:structure}). 

\subsection{Images Collection} \label{data:collect}
Our image acquisition system consists of two programmable cameras with the fixed focal length. The two cameras are positioned side by side, they will be calibrated before being used, and their locations remain unchanged during capturing. More importantly, the exposure time of the two cameras is set to 2ms and 10ms, respectively, so as to capture sharp images and the corresponding blurry ones.

We create and optimize a dual-camera shooting controller by using low-level programming language together with multi-thread technology, such a highly efficient controller allows our system to shoot both cameras simultaneously and further enables us to focus on fast-moving objects. In this work, we set the objects as cars only. Our intention is that car is one of the most commonly seen fast-moving objects in our daily life and it also plays an important role in computer vision, \textit{e.g.}, autopilot. Besides, the car moves horizontally with rotating wheels, which can provide different kinds of blur information, and at the same time, reduce interference from other factors, such as limb movement and deformation.

To ensure the diversity of the constructed dataset, we capture images in a variety of scenarios with different locations and times. We also try to cover as many kinds of cars as possible. Besides, images are mainly captured in the daytime, including morning, afternoon, and nightfall. Unlike RealBlur~\cite{rim2020real}, we do not attempt to capture images at night, because the low light conditions may damage the image quality, especially for fast-moving objects. 


\subsection{Data Cleaning} \label{data:clean}
As aforementioned, currently, our proposed dataset only includes moving cars, thus data cleaning mainly refers to detect, crop, and align cars in both the sharp images and the corresponding blurry images. 

As for the sharp images, detecting and cropping cars can be perfectly addressed by off-the-shelf state-of-the-art image segmentation models. In this work, we process the sharp images by using the Mask-RCNN~\cite{he2017mask}, pre-trained on the Microsoft COCO dataset~\cite{lin2014microsoft}. The detected bounding boxes of cars will be updated by non-maximum suppression, with the Intersection over Union (IoU) being set to 0.5. In order to guarantee the image quality of sharp cars, we filter the bounding boxes according to their importance. Specifically, we model the importance of a bounding box based on its prediction accuracy, location, and size. Only images with at least one qualified bounding box will be passed to the following process pipeline.

Dealing with blurry images is much more challenging than processing the above sharp images. The main reason is that most image segmentation models are trained with high-quality sharp images, making their performance unreliable on poor quality blurry images.
In light of such a generalization issue, we choose to design a heuristic algorithm to improve the quality of cropping and matching, by leveraging the Peak Signal-to-Noise Ratio (PSNR) metric and the sliding window technique. Specifically, for a given blurry image, after obtaining the approximate bounding box of the blurry car using Mask-RCNN, we define a candidate searching area, by setting its center point to the same position as the one of the initial bounding box, while its height and width to double. After that, a window frame, with the same size as the corresponding sharp car, will slide within the candidate searching area. For each sliding step, an image patch under the current window frame will be used to calculate the PSNR score towards the sharp car. In the end, the image patch with the highest PSNR score will be chosen as the blurry car image. Note that it is time-efficient yet reasonable to search within only the defined searching area instead of the whole image because the paired sharp and blurry images have already been roughly aligned as they are captured simultaneously. Besides, the whole matching process only involves similarity transformation, so that the blurry car remains natural without unexpected distortion. 

\subsection{Dataset Structure} \label{data:structure}
Using the image collection and data cleaning methods detailed as above, we propose a dataset with sharp and naturally blurry car pairs for motion deblurring, named Real Object Motion Blur (ROMB). In general, our ROMB dataset contains 13452 pairs of cars, captured in 16 scenarios. Image samples are shown in the experiment section. For the purpose of evaluation, we randomly select 10 scenarios as the training set and the rest 6 scenarios as the test set, resulting in 8533 pairs of training images and 4919 pairs of test images. Each scenario is independent in terms of date, time and location, making sure that there is no overlap between the training and, test set. 

\begin{table}[t]
\centering
\caption{Quantitative results of ablation study on ROMB. $\phi$ stands for the single branch baseline model, $\phi_{\text{CAT}}$, $\phi_{\text{ADD}}$, $\phi_{\text{ATT}}$ refer to model with concatenation, addition and attentive fusion method, respectively, and $\phi_{\text{EAL}}$ is $\phi$ with edge guided  loss.}
\label{tab:ablation_study}
\begin{tabularx}{0.48\textwidth}{@{\extracolsep{\fill}}lcccccc}
\toprule
Methods     & $\phi$  & $\phi_{\text{CAT}}$   & $\phi_{\text{ADD}}$   & $\phi_{\text{ATT}}$   & $\phi_{\text{EAL}}$   & EPAN           \\ \midrule\midrule
PSNR & 22.37 & 22.40 & 22.67 & 23.22 & 22.64 & \textbf{23.46} \\
SSIM & 0.813 & 0.818 & 0.824 & 0.834 & 0.825 & \textbf{0.846}      \\ \bottomrule
\end{tabularx}
\end{table}

\begin{figure*}[tbh]
    \centering
    \begin{subfigure}[b]{0.24\textwidth}
        \subfloat{\includegraphics[width=\textwidth]{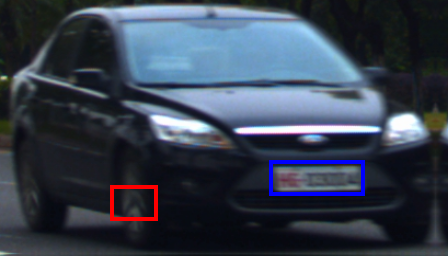}}
        \vspace*{0.2em}
        \subfloat{\includegraphics[width=0.325\textwidth]{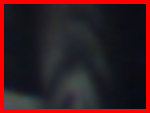}}
        \hspace*{0.1em}
        \subfloat{\includegraphics[width=0.64\textwidth]{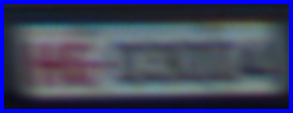}}
    \end{subfigure}
    \begin{subfigure}[b]{0.24\textwidth}
        \subfloat{\includegraphics[width=\textwidth]{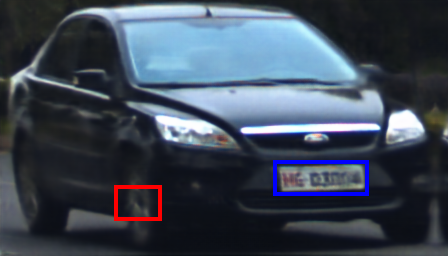}}
        \vspace*{0.2em}
        \subfloat{\includegraphics[width=0.325\textwidth]{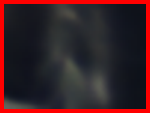}}
        \hspace*{0.1em}
        \subfloat{\includegraphics[width=0.64\textwidth]{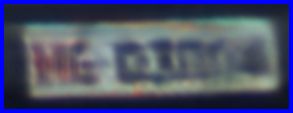}}
    \end{subfigure}
    \begin{subfigure}[b]{0.24\textwidth}
        \subfloat{\includegraphics[width=\textwidth]{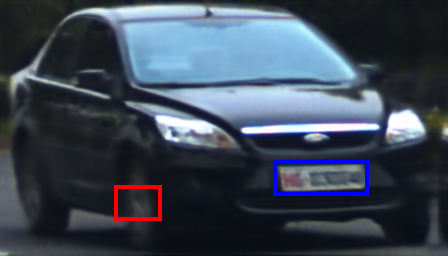}}
        \vspace*{0.2em}
        \subfloat{\includegraphics[width=0.325\textwidth]{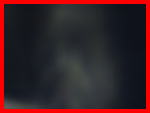}}
        \hspace*{0.1em}
        \subfloat{\includegraphics[width=0.64\textwidth]{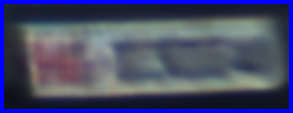}}
    \end{subfigure}
    \begin{subfigure}[b]{0.24\textwidth}
        \subfloat{\includegraphics[width=\textwidth]{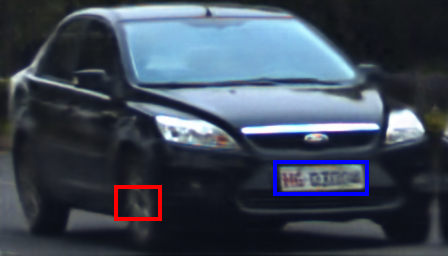}}
        \vspace*{0.2em}
        \subfloat{\includegraphics[width=0.325\textwidth]{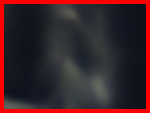}}
        \hspace*{0.1em}
        \subfloat{\includegraphics[width=0.64\textwidth]{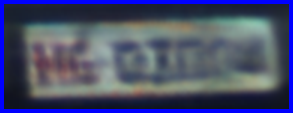}}
    \end{subfigure}

    \vspace*{0.2em}

    \begin{subfigure}[b]{0.24\textwidth}
        \subfloat{\includegraphics[width=\textwidth]{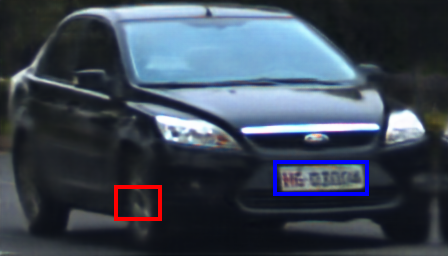}}
        \vspace*{0.2em}
        \subfloat{\includegraphics[width=0.325\textwidth]{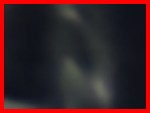}}
        \hspace*{0.1em}
        \subfloat{\includegraphics[width=0.64\textwidth]{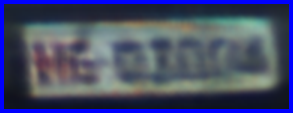}}
    \end{subfigure}
    \begin{subfigure}[b]{0.24\textwidth}
        \subfloat{\includegraphics[width=\textwidth]{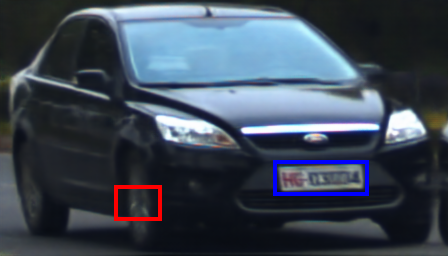}}
        \vspace*{0.2em}
        \subfloat{\includegraphics[width=0.325\textwidth]{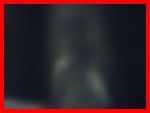}}
        \hspace*{0.1em}
        \subfloat{\includegraphics[width=0.64\textwidth]{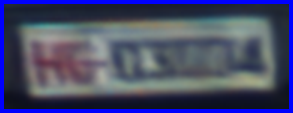}}
    \end{subfigure}
    \begin{subfigure}[b]{0.24\textwidth}
        \subfloat{\includegraphics[width=\textwidth]{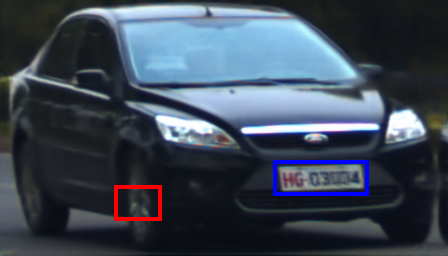}}
        \vspace*{0.2em}
        \subfloat{\includegraphics[width=0.325\textwidth]{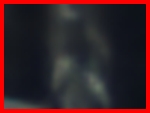}}
        \hspace*{0.1em}
        \subfloat{\includegraphics[width=0.64\textwidth]{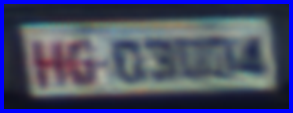}}
    \end{subfigure}
    \begin{subfigure}[b]{0.24\textwidth}
        \subfloat{\includegraphics[width=\textwidth]{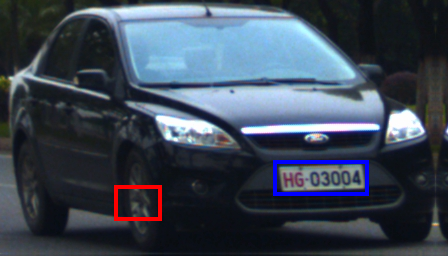}}
        \vspace*{0.2em}
        \subfloat{\includegraphics[width=0.325\textwidth]{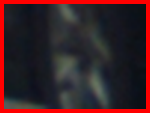}}
        \hspace*{0.1em}
        \subfloat{\includegraphics[width=0.64\textwidth]{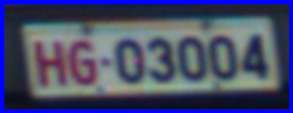}}
    \end{subfigure}
    
    \vspace*{0.4em}
    
    \begin{subfigure}[b]{0.24\textwidth}
        \subfloat{\includegraphics[width=\textwidth]{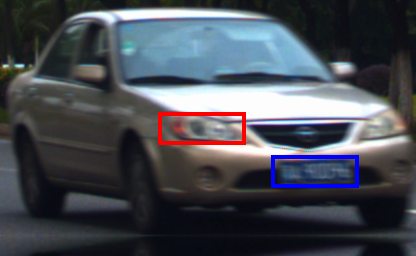}}
        \vspace*{0.2em}
        \subfloat{\includegraphics[width=0.48\textwidth]{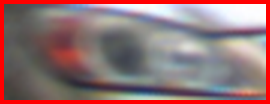}}
        \hspace*{0.1em}
        \subfloat{\includegraphics[width=0.48\textwidth]{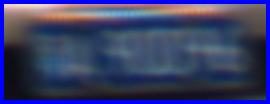}}
    \end{subfigure}
    \begin{subfigure}[b]{0.24\textwidth}
        \subfloat{\includegraphics[width=\textwidth]{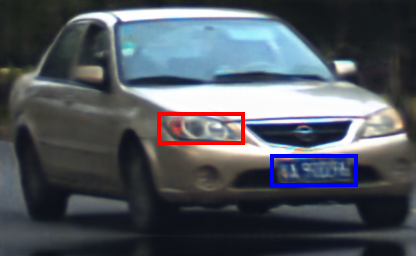}}
        \vspace*{0.2em}
        \subfloat{\includegraphics[width=0.48\textwidth]{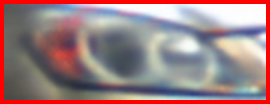}}
        \hspace*{0.1em}
        \subfloat{\includegraphics[width=0.48\textwidth]{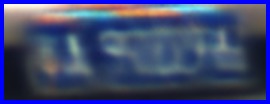}}
    \end{subfigure}
    \begin{subfigure}[b]{0.24\textwidth}
        \subfloat{\includegraphics[width=\textwidth]{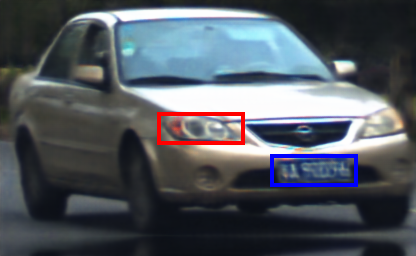}}
        \vspace*{0.2em}
        \subfloat{\includegraphics[width=0.48\textwidth]{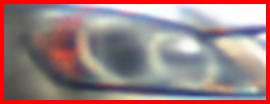}}
        \hspace*{0.1em}
        \subfloat{\includegraphics[width=0.48\textwidth]{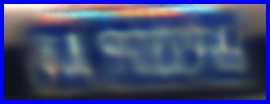}}
    \end{subfigure}
    \begin{subfigure}[b]{0.24\textwidth}
        \subfloat{\includegraphics[width=\textwidth]{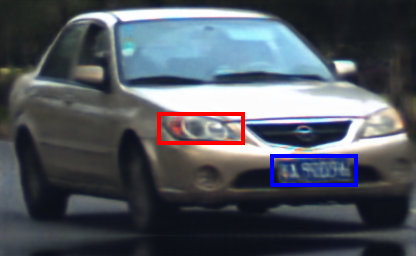}}
        \vspace*{0.2em}
        \subfloat{\includegraphics[width=0.48\textwidth]{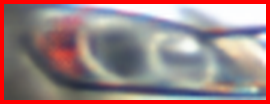}}
        \hspace*{0.1em}
        \subfloat{\includegraphics[width=0.48\textwidth]{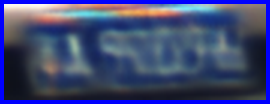}}
    \end{subfigure}

    \vspace*{0.2em}

    \begin{subfigure}[b]{0.24\textwidth}
        \subfloat{\includegraphics[width=\textwidth]{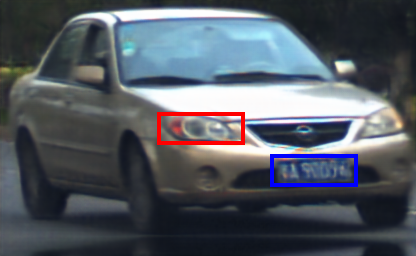}}
        \vspace*{0.2em}
        \subfloat{\includegraphics[width=0.48\textwidth]{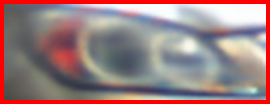}}
        \hspace*{0.1em}
        \subfloat{\includegraphics[width=0.48\textwidth]{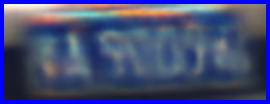}}
    \end{subfigure}
    \begin{subfigure}[b]{0.24\textwidth}
        \subfloat{\includegraphics[width=\textwidth]{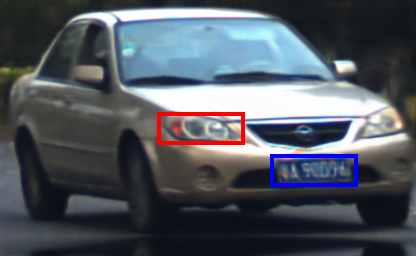}}
        \vspace*{0.2em}
        \subfloat{\includegraphics[width=0.48\textwidth]{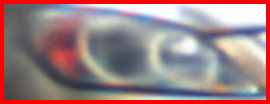}}
        \hspace*{0.1em}
        \subfloat{\includegraphics[width=0.48\textwidth]{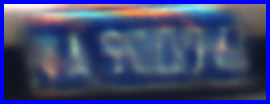}}
    \end{subfigure}
    \begin{subfigure}[b]{0.24\textwidth}
        \subfloat{\includegraphics[width=\textwidth]{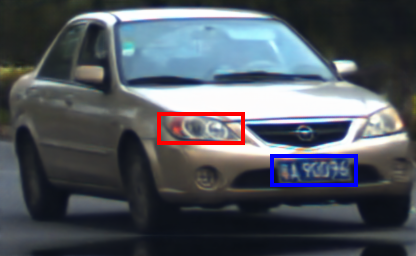}}
        \vspace*{0.2em}
        \subfloat{\includegraphics[width=0.48\textwidth]{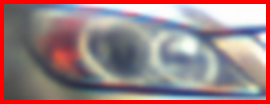}}
        \hspace*{0.1em}
        \subfloat{\includegraphics[width=0.48\textwidth]{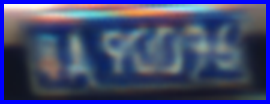}}
    \end{subfigure}
    \begin{subfigure}[b]{0.24\textwidth}
        \subfloat{\includegraphics[width=\textwidth]{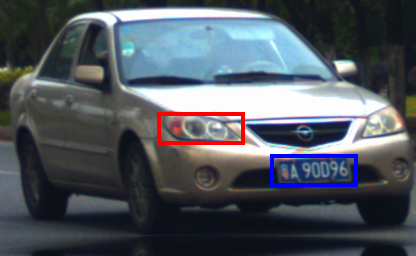}}
        \vspace*{0.2em}
        \subfloat{\includegraphics[width=0.48\textwidth]{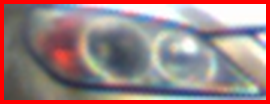}}
        \hspace*{0.1em}
        \subfloat{\includegraphics[width=0.48\textwidth]{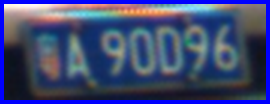}}
    \end{subfigure}

    \caption{Qualitative results of SOTA comparisons on ROMB dataset. 
    For each scene, images of the first row from left to right are 
    blurry image, SRN~\cite{tao2018scale}, DeblurGAN-v2~\cite{kupyn2019deblurgan}, Gao \textit{et al}.~\cite{gao2019dynamic}, 
    while the second row are DBGAN~\cite{zhang2020deblurring}, Suin \textit{et al}.~\cite{suin2020spatially}, EPAN(Ours) and sharp image, respectively. Two zoom-in patches are presented below each image. (Best viewed in high-resolution color)} \label{fig:carvision_dataset}
\end{figure*}


\begin{table*}[t]
\centering
\caption{Quantitative results of SOTA comparisons on ROMB dataset. Best results are highlighted in bold.}
\label{tab:ROMB}
\begin{tabularx}{0.98\textwidth}{@{\extracolsep{\fill}}lccc cc c}
\toprule
Methods     
& SRN~\cite{tao2018scale} &DeblurGAN-v2~\cite{kupyn2019deblurgan} 
& Gao \textit{et al}.~\cite{gao2019dynamic} 
& DBGAN~\cite{zhang2020deblurring}  & Suin \textit{et al}.~\cite{suin2020spatially} 
 & EPAN (Ours)         \\ \midrule\midrule
PSNR &21.87 &21.57 &21.91 &22.16 &22.64 &\textbf{23.46}  \\
SSIM &0.821 &0.804 &0.826 &0.828 &0.825 &\textbf{0.846}               \\ \bottomrule
\end{tabularx}
\end{table*}

\begin{table*}[t]
\centering
\caption{Quantitative results of SOTA comparisons on RealBlur-J dataset. Best results are highlighted in bold.}
\label{tab:RealBlur}
\begin{tabularx}{0.99\textwidth}{@{\extracolsep{\fill}}l cc cccc cc}
\toprule
    Method 
    & Xu~\textit{et al}.~\cite{xu2013unnatural} & Pan~\textit{et al}.~\cite{pan2016blind}  & Nah ~\textit{et al.}~\cite{nah2017deep} & Zhang~\textit{et al}.~\cite{zhang2018dynamic}  & SRN~\cite{tao2018scale} & DeblurGAN-v2~\cite{kupyn2019deblurgan} & DMPHN~\cite{zhang2019deep} & EPAN (Ours)   \\
    \midrule\midrule
    PSNR & 27.14 & 27.22  & 27.87   & 27.80 & 28.56 & 28.70 & 28.42 & \textbf{28.80}  \\
    SSIM & 0.830 & 0.790  & 0.827   & 0.847 & 0.867 & 0.866 & 0.860 & \textbf{0.879}  \\
    \bottomrule
\end{tabularx}
\end{table*}

\begin{table*}[t]
\centering
\caption{Quantitative results of SOTA comparisons on GoPro dataset. Best results are highlighted in bold, and the second best results are highlighted with bracket.}
\label{tab:GoPro}
\begin{tabularx}{0.99\textwidth}{@{\extracolsep{\fill}}l ccc ccc cc}
\toprule
    Method 
    &SRN~\cite{tao2018scale} &DeblurGAN-v2~\cite{kupyn2019deblurgan} &Gao~\textit{et al}.~\cite{gao2019dynamic} &DMPHN~\cite{zhang2019deep} &DBGAN~\cite{zhang2020deblurring} 
    &Suin~\textit{et al}.~\cite{suin2020spatially} &MTRNN~\cite{park2020multi}
    & EPAN (Ours) \\
    \midrule\midrule
    PSNR & 30.26 & 29.55 & 30.90 & 31.20 & 31.10 & \textbf{31.85} & 31.15 & [31.42] \\
    SSIM & 0.934 & 0.934 & 0.935 & 0.940 & 0.940 & [0.948] & 0.945 & \textbf{0.964} \\
    \bottomrule
\end{tabularx}
\end{table*}

\begin{figure*}[tbh]
    \centering
    \begin{subfigure}[h]{0.192\textwidth}
        \subfloat{\includegraphics[width=\textwidth]{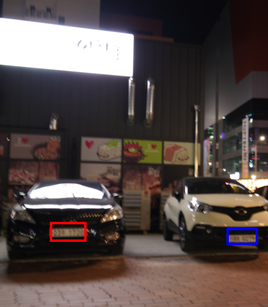}}
        \vspace*{-1.1em}
        \subfloat{\includegraphics[width=0.48\textwidth]{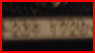}}
        \hspace*{0.1em}
        \subfloat{\includegraphics[width=0.48\textwidth]{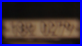}}
    \end{subfigure}
    \begin{subfigure}[h]{0.192\textwidth}
        \subfloat{\includegraphics[width=\textwidth]{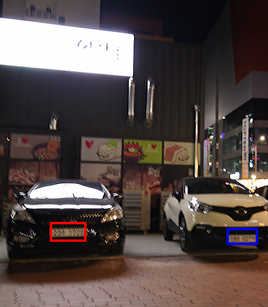}}
        \vspace*{-1.1em}
        \subfloat{\includegraphics[width=0.48\textwidth]{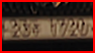}}
        \hspace*{0.1em}
        \subfloat{\includegraphics[width=0.48\textwidth]{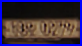}}
    \end{subfigure}
    \begin{subfigure}[h]{0.192\textwidth}
        \subfloat{\includegraphics[width=\textwidth]{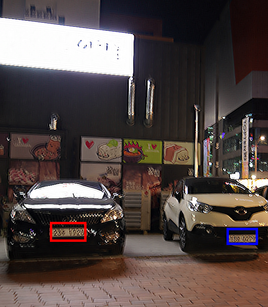}}
        \vspace*{-1.1em}
        \subfloat{\includegraphics[width=0.48\textwidth]{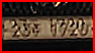}}
        \hspace*{0.1em}
        \subfloat{\includegraphics[width=0.48\textwidth]{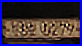}}
    \end{subfigure}
    \begin{subfigure}[h]{0.192\textwidth}
        \subfloat{\includegraphics[width=\textwidth]{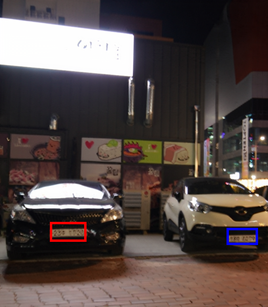}}
        \vspace*{-1.1em}
        \subfloat{\includegraphics[width=0.48\textwidth]{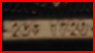}}
        \hspace*{0.1em}
        \subfloat{\includegraphics[width=0.48\textwidth]{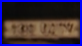}}
    \end{subfigure}
    \begin{subfigure}[h]{0.192\textwidth}
        \subfloat{\includegraphics[width=\textwidth]{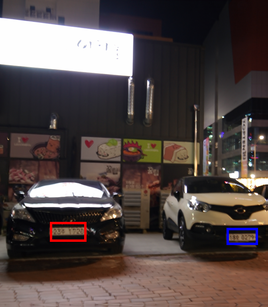}}
        \vspace*{-1.1em}
        \subfloat{\includegraphics[width=0.48\textwidth]{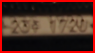}}
        \hspace*{0.1em}
        \subfloat{\includegraphics[width=0.48\textwidth]{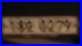}}
    \end{subfigure}

    \vspace*{0.1em}

    \begin{subfigure}[h]{0.192\textwidth}
        \subfloat{\includegraphics[width=\textwidth]{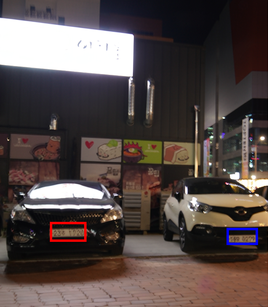}}
        \vspace*{-1.1em}
        \subfloat{\includegraphics[width=0.48\textwidth]{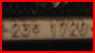}}
        \hspace*{0.1em}
        \subfloat{\includegraphics[width=0.48\textwidth]{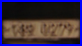}}
    \end{subfigure}
    \begin{subfigure}[h]{0.192\textwidth}
        \subfloat{\includegraphics[width=\textwidth]{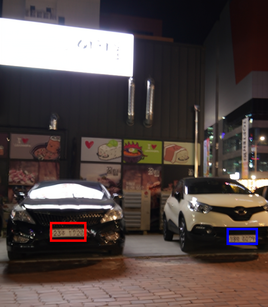}}
        \vspace*{-1.1em}
        \subfloat{\includegraphics[width=0.48\textwidth]{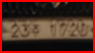}}
        \hspace*{0.1em}
        \subfloat{\includegraphics[width=0.48\textwidth]{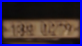}}
    \end{subfigure}
    \begin{subfigure}[h]{0.192\textwidth}
        \subfloat{\includegraphics[width=\textwidth]{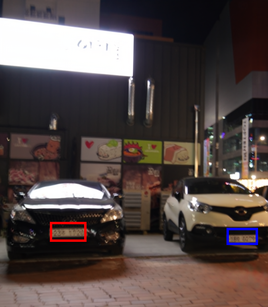}}
        \vspace*{-1.1em}
        \subfloat{\includegraphics[width=0.48\textwidth]{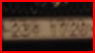}}
        \hspace*{0.1em}
        \subfloat{\includegraphics[width=0.48\textwidth]{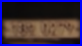}}
    \end{subfigure}
    \begin{subfigure}[h]{0.192\textwidth}
        \subfloat{\includegraphics[width=\textwidth]{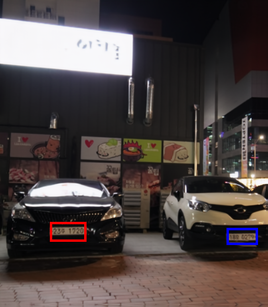}}
        \vspace*{-1.1em}
        \subfloat{\includegraphics[width=0.48\textwidth]{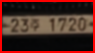}}
        \hspace*{0.1em}
        \subfloat{\includegraphics[width=0.48\textwidth]{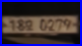}}
    \end{subfigure}
    \begin{subfigure}[h]{0.192\textwidth}
        \subfloat{\includegraphics[width=\textwidth]{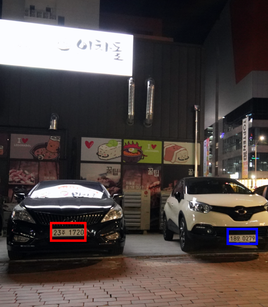}}
        \vspace*{-1.1em}
        \subfloat{\includegraphics[width=0.48\textwidth]{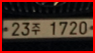}}
        \hspace*{0.1em}
        \subfloat{\includegraphics[width=0.48\textwidth]{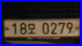}}
    \end{subfigure}
    
    \vspace*{0.4em}
    
    \begin{subfigure}[h]{0.192\textwidth}
        \subfloat{\includegraphics[width=\textwidth]{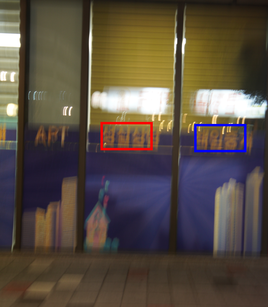}}
        \vspace*{-1.1em}
        \subfloat{\includegraphics[width=0.48\textwidth]{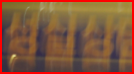}}
        \hspace*{0.1em}
        \subfloat{\includegraphics[width=0.48\textwidth]{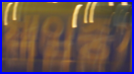}}
    \end{subfigure}
    \begin{subfigure}[h]{0.192\textwidth}
        \subfloat{\includegraphics[width=\textwidth]{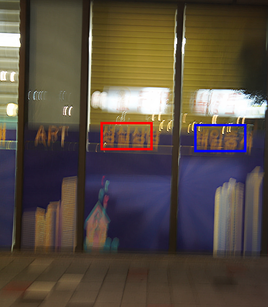}}
        \vspace*{-1.1em}
        \subfloat{\includegraphics[width=0.48\textwidth]{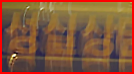}}
        \hspace*{0.1em}
        \subfloat{\includegraphics[width=0.48\textwidth]{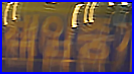}}
    \end{subfigure}
    \begin{subfigure}[h]{0.192\textwidth}
        \subfloat{\includegraphics[width=\textwidth]{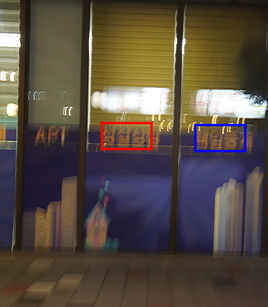}}
        \vspace*{-1.1em}
        \subfloat{\includegraphics[width=0.48\textwidth]{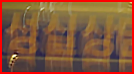}}
        \hspace*{0.1em}
        \subfloat{\includegraphics[width=0.48\textwidth]{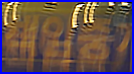}}
    \end{subfigure}
    \begin{subfigure}[h]{0.192\textwidth}
        \subfloat{\includegraphics[width=\textwidth]{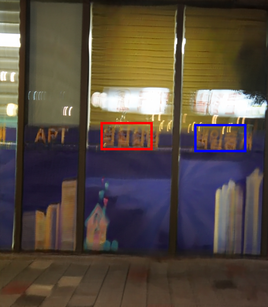}}
        \vspace*{-1.1em}
        \subfloat{\includegraphics[width=0.48\textwidth]{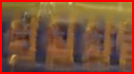}}
        \hspace*{0.1em}
        \subfloat{\includegraphics[width=0.48\textwidth]{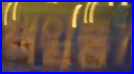}}
    \end{subfigure}
    \begin{subfigure}[h]{0.192\textwidth}
        \subfloat{\includegraphics[width=\textwidth]{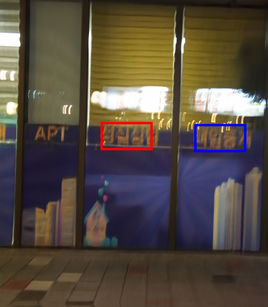}}
        \vspace*{-1.1em}
        \subfloat{\includegraphics[width=0.48\textwidth]{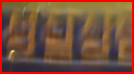}}
        \hspace*{0.1em}
        \subfloat{\includegraphics[width=0.48\textwidth]{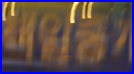}}
    \end{subfigure}

    \vspace*{0.1em}

    \begin{subfigure}[h]{0.192\textwidth}
        \subfloat{\includegraphics[width=\textwidth]{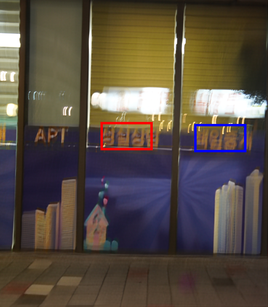}}
        \vspace*{-1.1em}
        \subfloat{\includegraphics[width=0.48\textwidth]{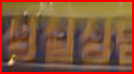}}
        \hspace*{0.1em}
        \subfloat{\includegraphics[width=0.48\textwidth]{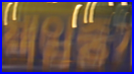}}
    \end{subfigure}
    \begin{subfigure}[h]{0.192\textwidth}
        \subfloat{\includegraphics[width=\textwidth]{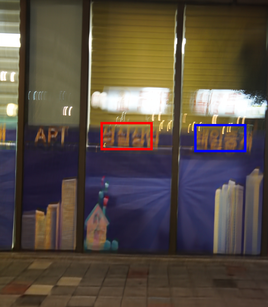}}
        \vspace*{-1.1em}
        \subfloat{\includegraphics[width=0.48\textwidth]{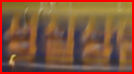}}
        \hspace*{0.1em}
        \subfloat{\includegraphics[width=0.48\textwidth]{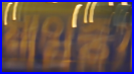}}
    \end{subfigure}
    \begin{subfigure}[h]{0.192\textwidth}
        \subfloat{\includegraphics[width=\textwidth]{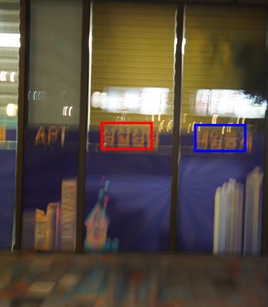}}
        \vspace*{-1.1em}
        \subfloat{\includegraphics[width=0.48\textwidth]{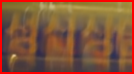}}
        \hspace*{0.1em}
        \subfloat{\includegraphics[width=0.48\textwidth]{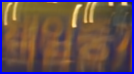}}
    \end{subfigure}
    \begin{subfigure}[h]{0.192\textwidth}
        \subfloat{\includegraphics[width=\textwidth]{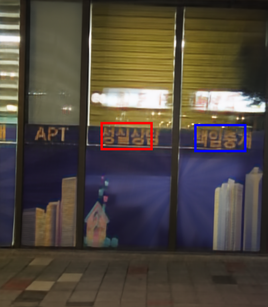}}
        \vspace*{-1.1em}
        \subfloat{\includegraphics[width=0.48\textwidth]{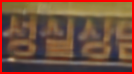}}
        \hspace*{0.1em}
        \subfloat{\includegraphics[width=0.48\textwidth]{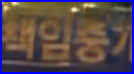}}
    \end{subfigure}
    \begin{subfigure}[h]{0.192\textwidth}
        \subfloat{\includegraphics[width=\textwidth]{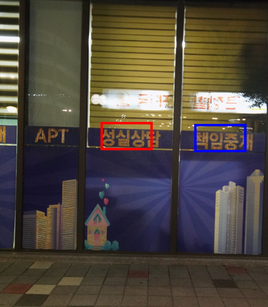}}
        \vspace*{-1.1em}
        \subfloat{\includegraphics[width=0.48\textwidth]{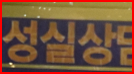}}
        \hspace*{0.1em}
        \subfloat{\includegraphics[width=0.48\textwidth]{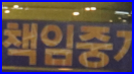}}
    \end{subfigure}

    \caption{Qualitative results of SOTA comparisons on RealBlur-J dataset. 
    For each scene, images of the first row from left to right are blurry image,
    Xu~\textit{et al}.~\cite{xu2013unnatural}, Pan~\textit{et al}.~\cite{pan2016blind}, Nah ~\textit{et al.}~\cite{nah2017deep}, Zhang~\textit{et al}.~\cite{zhang2018dynamic},
    while the second row are
    SRN~\cite{tao2018scale}, DeblurGAN-v2~\cite{kupyn2019deblurgan}, DMPHN~\cite{zhang2019deep} EPAN(Ours) and sharp image, respectively. Two zoom-in patches are presented below each image. (Best viewed in high-resolution color)
    } \label{fig:realblurj_dataset}
\end{figure*}

\begin{figure*}[tbh]
    \centering
    \begin{subfigure}[b]{0.192\textwidth}
        \subfloat{\includegraphics[width=\textwidth]{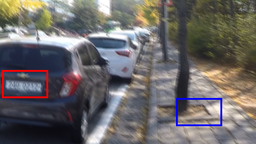}}
        \vspace*{0.2em}
        \subfloat{\includegraphics[width=0.48\textwidth]{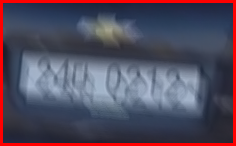}}
        \hspace*{0.1em}
        \subfloat{\includegraphics[width=0.48\textwidth]{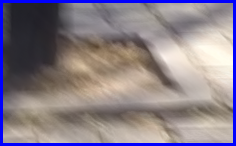}}
    \end{subfigure}
    \begin{subfigure}[b]{0.192\textwidth}
        \subfloat{\includegraphics[width=\textwidth]{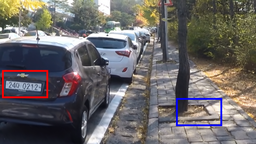}}
        \vspace*{0.2em}
        \subfloat{\includegraphics[width=0.48\textwidth]{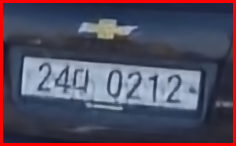}}
        \hspace*{0.1em}
        \subfloat{\includegraphics[width=0.48\textwidth]{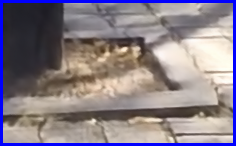}}
    \end{subfigure}
    \begin{subfigure}[b]{0.192\textwidth}
        \subfloat{\includegraphics[width=\textwidth]{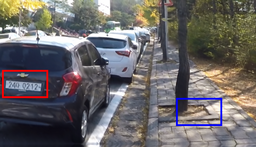}}
        \vspace*{0.2em}
        \subfloat{\includegraphics[width=0.48\textwidth]{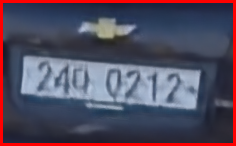}}
        \hspace*{0.1em}
        \subfloat{\includegraphics[width=0.48\textwidth]{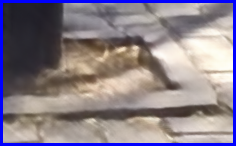}}
    \end{subfigure}
    \begin{subfigure}[b]{0.192\textwidth}
        \subfloat{\includegraphics[width=\textwidth]{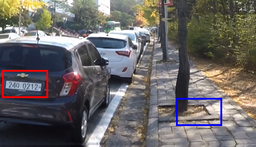}}
        \vspace*{0.2em}
        \subfloat{\includegraphics[width=0.48\textwidth]{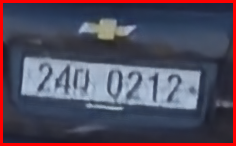}}
        \hspace*{0.1em}
        \subfloat{\includegraphics[width=0.48\textwidth]{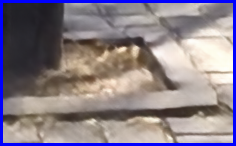}}
    \end{subfigure}
    \begin{subfigure}[b]{0.192\textwidth}
        \subfloat{\includegraphics[width=\textwidth]{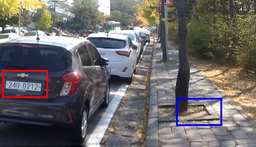}}
        \vspace*{0.2em}
        \subfloat{\includegraphics[width=0.48\textwidth]{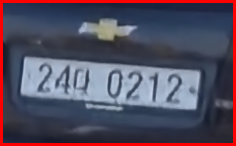}}
        \hspace*{0.1em}
        \subfloat{\includegraphics[width=0.48\textwidth]{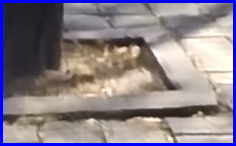}}
    \end{subfigure}

    \vspace*{0.2em}

    \begin{subfigure}[b]{0.192\textwidth}
        \subfloat{\includegraphics[width=\textwidth]{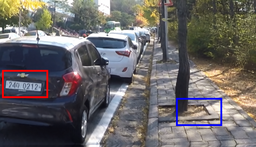}}
        \vspace*{0.2em}
        \subfloat{\includegraphics[width=0.48\textwidth]{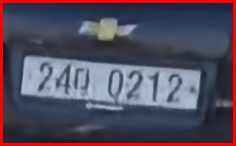}}
        \hspace*{0.1em}
        \subfloat{\includegraphics[width=0.48\textwidth]{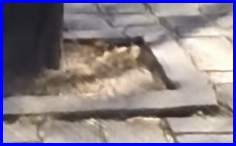}}
    \end{subfigure}
    \begin{subfigure}[b]{0.192\textwidth}
        \subfloat{\includegraphics[width=\textwidth]{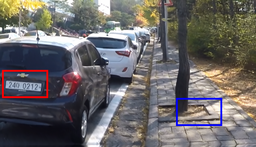}}
        \vspace*{0.2em}
        \subfloat{\includegraphics[width=0.48\textwidth]{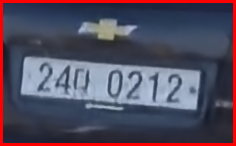}}
        \hspace*{0.1em}
        \subfloat{\includegraphics[width=0.48\textwidth]{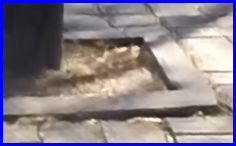}}
    \end{subfigure}
    \begin{subfigure}[b]{0.192\textwidth}
        \subfloat{\includegraphics[width=\textwidth]{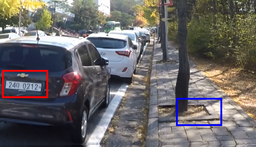}}
        \vspace*{0.2em}
        \subfloat{\includegraphics[width=0.48\textwidth]{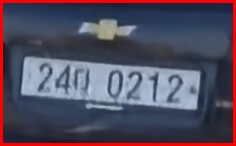}}
        \hspace*{0.1em}
        \subfloat{\includegraphics[width=0.48\textwidth]{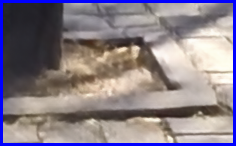}}
    \end{subfigure}
    \begin{subfigure}[b]{0.192\textwidth}
        \subfloat{\includegraphics[width=\textwidth]{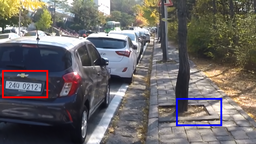}}
        \vspace*{0.2em}
        \subfloat{\includegraphics[width=0.48\textwidth]{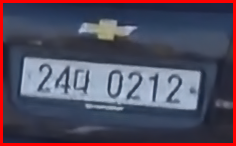}}
        \hspace*{0.1em}
        \subfloat{\includegraphics[width=0.48\textwidth]{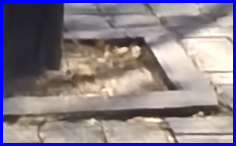}}
    \end{subfigure}
    \begin{subfigure}[b]{0.192\textwidth}
        \subfloat{\includegraphics[width=\textwidth]{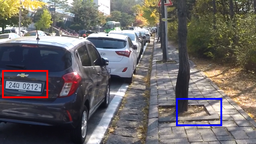}}
        \vspace*{0.2em}
        \subfloat{\includegraphics[width=0.48\textwidth]{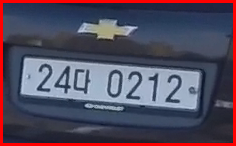}}
        \hspace*{0.1em}
        \subfloat{\includegraphics[width=0.48\textwidth]{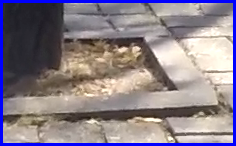}}
    \end{subfigure}
    
    \vspace*{0.4em}
    
    \begin{subfigure}[b]{0.192\textwidth}
        \subfloat{\includegraphics[width=\textwidth]{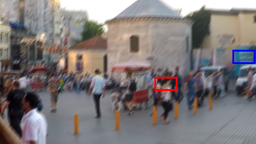}}
        \vspace*{0.2em}
        \subfloat{\includegraphics[width=0.48\textwidth]{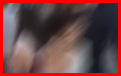}}
        \hspace*{0.1em}
        \subfloat{\includegraphics[width=0.48\textwidth]{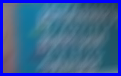}}
    \end{subfigure}
    \begin{subfigure}[b]{0.192\textwidth}
        \subfloat{\includegraphics[width=\textwidth]{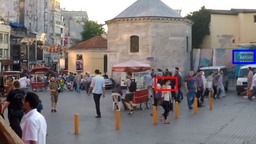}}
        \vspace*{0.2em}
        \subfloat{\includegraphics[width=0.48\textwidth]{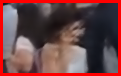}}
        \hspace*{0.1em}
        \subfloat{\includegraphics[width=0.48\textwidth]{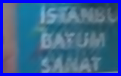}}
    \end{subfigure}
    \begin{subfigure}[b]{0.192\textwidth}
        \subfloat{\includegraphics[width=\textwidth]{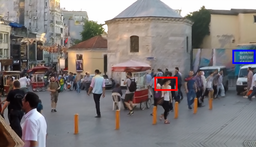}}
        \vspace*{0.2em}
        \subfloat{\includegraphics[width=0.48\textwidth]{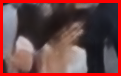}}
        \hspace*{0.1em}
        \subfloat{\includegraphics[width=0.48\textwidth]{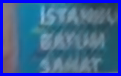}}
    \end{subfigure}
    \begin{subfigure}[b]{0.192\textwidth}
        \subfloat{\includegraphics[width=\textwidth]{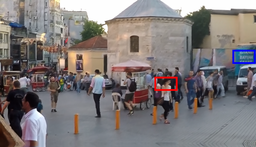}}
        \vspace*{0.2em}
        \subfloat{\includegraphics[width=0.48\textwidth]{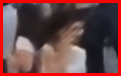}}
        \hspace*{0.1em}
        \subfloat{\includegraphics[width=0.48\textwidth]{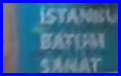}}
    \end{subfigure}
    \begin{subfigure}[b]{0.192\textwidth}
        \subfloat{\includegraphics[width=\textwidth]{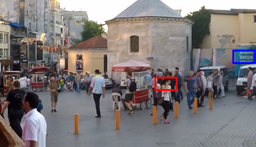}}
        \vspace*{0.2em}
        \subfloat{\includegraphics[width=0.48\textwidth]{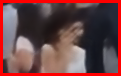}}
        \hspace*{0.1em}
        \subfloat{\includegraphics[width=0.48\textwidth]{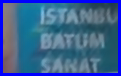}}
    \end{subfigure}

    \vspace*{0.2em}

    \begin{subfigure}[b]{0.192\textwidth}
        \subfloat{\includegraphics[width=\textwidth]{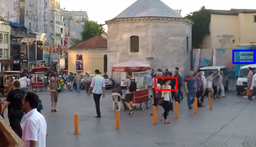}}
        \vspace*{0.2em}
        \subfloat{\includegraphics[width=0.48\textwidth]{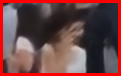}}
        \hspace*{0.1em}
        \subfloat{\includegraphics[width=0.48\textwidth]{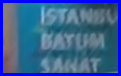}}
    \end{subfigure}
    \begin{subfigure}[b]{0.192\textwidth}
        \subfloat{\includegraphics[width=\textwidth]{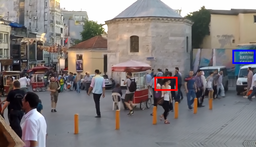}}
        \vspace*{0.2em}
        \subfloat{\includegraphics[width=0.48\textwidth]{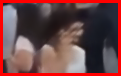}}
        \hspace*{0.1em}
        \subfloat{\includegraphics[width=0.48\textwidth]{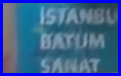}}
    \end{subfigure}
    \begin{subfigure}[b]{0.192\textwidth}
        \subfloat{\includegraphics[width=\textwidth]{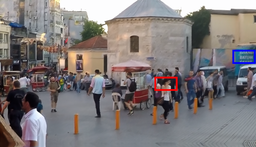}}
        \vspace*{0.2em}
        \subfloat{\includegraphics[width=0.48\textwidth]{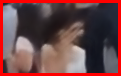}}
        \hspace*{0.1em}
        \subfloat{\includegraphics[width=0.48\textwidth]{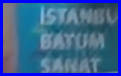}}
    \end{subfigure}
    \begin{subfigure}[b]{0.192\textwidth}
        \subfloat{\includegraphics[width=\textwidth]{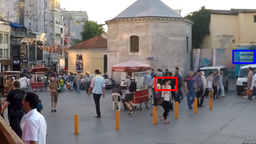}}
        \vspace*{0.2em}
        \subfloat{\includegraphics[width=0.48\textwidth]{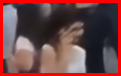}}
        \hspace*{0.1em}
        \subfloat{\includegraphics[width=0.48\textwidth]{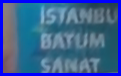}}
    \end{subfigure}
    \begin{subfigure}[b]{0.192\textwidth}
        \subfloat{\includegraphics[width=\textwidth]{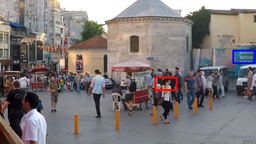}}
        \vspace*{0.2em}
        \subfloat{\includegraphics[width=0.48\textwidth]{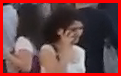}}
        \hspace*{0.1em}
        \subfloat{\includegraphics[width=0.48\textwidth]{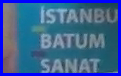}}
    \end{subfigure}

    \caption{Qualitative results of SOTA comparisons on GoPro dataset. 
    For each scene, images of the first row from left to right are blurry image, SRN~\cite{tao2018scale}, DeblurGAN-v2~\cite{kupyn2019deblurgan}, Gao~\textit{et al.}~\cite{gao2019dynamic}, DMPHN~\cite{zhang2019deep},
    while the second row are DBGAN~\cite{zhang2020deblurring}, Suin~\textit{et al.}~\cite{suin2020spatially}, MTRNN~\cite{park2020multi}, EPAN(Ours) and sharp image, respectively. Two zoom-in patches are presented below each image. (Best viewed in high-resolution color)} \label{fig:gopro_dataset}
\end{figure*}

\section{Experiments} \label{experiments}

\subsection{Settings} \label{exp:setting}

\noindent\textbf{Edge Detection.}
EPAN requires edge maps of both blurry and sharp cars in the training phase. In order to find the most suitable edge detector, we have conducted a thorough study on existing related algorithms, including the Canny edge detector~\cite{canny1986computational}, DeepEdge~\cite{bertasius2015deepedge}, HED~\cite{xie2015holistically} and CASENet~\cite{yu2017casenet}. 
Compared to those detected by the counterparts, edges detected by HED are found to contain much richer details and smoother edges, with fewer overlaps and artifacts, 
therefore we choose HED in our experiments.
Besides, HED also has a rather simple network structure, resulting in faster running speed. Although EPAN also needs to extract edge maps for the given blurry images during testing, the whole detection process costs within milliseconds by using HED and it can also be done offline. Note that only the enhanced feature of the blurry edge will be used to augment the deblurring in testing, therefore it will not cause huge harm to the final output even when the detected edge contains minor errors.

\noindent\textbf{Implementation Details.} 
We validate the effectiveness of each component of EPAN with a series of ablation studies conducted on ROMB. Besides, we also compare EPAN with existing state-of-the-art motion deblurring models on datasets ROMB,  RealBlur~\cite{rim2020real} and GoPro~\cite{nah2017deep}. Results and analyses regarding each dataset will be given in the following related subsections. As a general approach~\cite{nah2017deep, rim2020real}, we measure and report the quantitative results with both PSNR and SSIM~\cite{wang2004image} metrics.

For each dataset, in order to alleviate overfitting issues, several data augmentation techniques are employed. 
Specifically, an image patch with a specific size, \textit{i.e.}, $640 \times 360$ for both ROMB and GoPro, $512 \times 576$ for RealBlur, will be randomly cropped from the input image loaded with the original size, before being fed into the model. The image will be further randomly flipped and rotated. Note that all the above operations should be performed in exactly the same way in blurry and sharp images and edge maps. Besides, the input queue will be randomly rearranged at the beginning of each training epoch. 

In all experiements, we set $\lambda_C = 4$ in Eq.~\eqref{eq:cdn} and $\lambda_E = 4$ in Eq.~\eqref{eq:een}. We implement EPAN with PyTorch~\cite{NEURIPS2019_9015}, and set the training batch size to 4.
All network components are randomly initialized. Adam optimizer is used to optimize the training, with coefficients being set to $\beta_1=0.9, \beta_2=0.999, \epsilon=1e^{-7}$. We initialize the learning rate as $1e^{-3}$, and exponentially decay it with power 0.3 to $1e^{-6}$ in 1500 epochs. The whole training procedure takes around 72 hours 
on a GeForce GTX 1080 Ti GPU. 


\subsection{Ablation Study}
To verify the effectiveness of the critical components of EPAN, we design a series of ablation studies, trained and tested on the proposed ROMB dataset. 
Specifically, our baseline model contains only the CDN branch (the topper branch of Fig.~\ref{fig:model_archi}) and the MSE loss function (see Eq.~\eqref{eq:mse}), denoted as $\phi$. To further verify the effectiveness of attentive fusion modules, we compare them with two other simple counterparts, namely concatenation and addition. Here we design three models, including $\phi_{\text{ATT}}$ (EPAN without edge guided  loss), $\phi_{\text{CAT}}$ (replace attentive fusion modules with feature concatenations in $\phi_{\text{ATT}}$) and $\phi_{\text{ADD}}$ (replace attentive fusion modules with feature addition in $\phi_{\text{ATT}}$).
Besides, to locate the performance of the edge-guided loss function, 
we test the combination of baseline $\phi$ and edge-guided loss (EPAN without EEN),
denoted as $\phi_{\text{EAL}}$. For fair comparisons, all the above models are examined under the same experiment settings as EPAN.

As presented in TABLE~\ref{tab:ablation_study}, compared to $\phi$, model $\phi_{\text{CAT}}$,  $\phi_{\text{ADD}}$ and $\phi_{\text{ATT}}$ all see improvement, indicating that using edge prior knowledge with an auxiliary branch can benefit motion deblurring, no matter which kind of fusion method is employed. Meanwhile, $\phi_{\text{ATT}}$ outperforms both $\phi_{\text{CAT}}$ and $\phi_{\text{ADD}}$, validating the superiority of attentive fusion mechanism over naive concatenation and addition. The main reason is that edge-related features mainly contain high-frequency information, treating them as spatial attention masks and fusing them through multiplication will help the content branch highlight the edge area, leading to results with better geometric structure. While by comparing $\phi_{\text{EAL}}$ with $\phi$, we can see that adding the edge-guided  loss function can assist networks to converge to a better state. It is because that edge guided  loss can guide the network to target more on the high-frequency area, which is beneficial for improving image sharpness. Finally, by using both attentive fusion mechanism and edge-guided loss function, our EPAN gets the best performance. It is clear that EPAN makes good use of edge prior knowledge in both network structure and training optimization, therefore it surpasses all other ablation models with a healthy margin.

\subsection{Comparison on ROMB}
We compare EPAN with existing SOTA single image motion deblurring models, including SRN~\cite{tao2018scale}, DeblurGAN-v2~\cite{kupyn2019deblurgan}, Gao \textit{et al}.~\cite{gao2019dynamic}, DBGAN~\cite{zhang2020deblurring} and Suin \textit{et al}.~\cite{suin2020spatially}.
Codes for the first three models are obtained from the officially released, as for DBGAN and Suin \textit{et al}., we cannot find their implementations and therefore choose to replicate them at our best.
All models are trained on the ROMB training set from scratch, with hyperparameters being set to the values suggested in corresponding papers. We then evaluate all models on the ROMB test set after training converged and show the visual results generated by all models, input blurry images, and ground truth sharp images in Fig.~\ref{fig:carvision_dataset}. 

It can be seen that all models succeed in improving the image quality to some extends, while images processed by EPAN achieve the best visual quality. As presented in those specific zoom-in closer view patches, EPAN can restore more detailed information, especially in those regions with more edges, \textit{e.g.}, the car license plate. This is mainly because that EPAN embeds the edge prior knowledge into both its model structure and training constraint, while other models do not. Besides, quantitative results for all models are given in TABLE~\ref{tab:ROMB}. As illustrated, the quantitative performances are consistent with the qualitative ones, specifically, the values of the different models are quite close and EPAN achieves the best.

\subsection{Comparison on RealBlur}
RealBlur dataset~\cite{rim2020real} contains 4738 pairs of blurry and sharp images, with 3758 pairs being the training set and the rest being the test set. Although the dataset provides two subsets, including RealBlur-R in camera raw format and RealBlur-J in JPEG format, we discard the former for two main reasons. Firstly, RealBlur-R contains lots of images that are too dark to recognize visual contents, not to mention extracting edges from them. Secondly, in order to use RealBlur-R, it is required to apply
a handful of post-processing operations, \textit{e.g.}, white balance, demosaicing, and denoising, we worry that such operations may alter the image contents, making the blurry images not as natural as the origin. Therefore, for the sake of fairness, we use only the RealBlur-J subset in the following experiments. 

We evaluate our EPAN following the same settings as detailed in \cite{rim2020real} and compare it with several SOTA methods. The quantitative results are given in TABLE~\ref{tab:RealBlur}. It is clear that traditional optimization-based methods, \textit{i.e.}, Xu~\textit{et al}.~\cite{xu2013unnatural} and Pan~\textit{et al}.~\cite{pan2016blind}, do not perform well on such naturally captured dataset, mainly because that the motion kernels are difficult to estimate correctly. All deep learning-based models achieve comparable performances, with EPAN being the best. Such improvements are brought by leveraging the edge prior knowledge, which is very helpful in handling regions with more edges, especially the text contents, as can be seen in the visual results in Fig.~\ref{fig:realblurj_dataset}.






\subsection{Comparison on GoPro}
The GoPro dataset~\cite{nah2017deep} provides paired blurry and sharp images, where the former is synthesized by averaging the latter. And it has 2103 image pairs in the training set and 1111 pairs in the test set. Qualitative and quantitative results are given in Fig.~\ref{fig:gopro_dataset} and TABLE~\ref{tab:GoPro}, respectively. 

With reference to those zoom-in patches in Fig.~\ref{fig:gopro_dataset}, we can see that our EPAN outperforms all other models in handling regions with more edges, i.e., the text contents, the pavement with bricks, and the crowds. This is consistent with results from ROMB and RealBlur dataset, and it shows that EPAN can restore more detailed information by embedding the edge prior knowledge. Besides, as illustrated in TABLE~\ref{tab:GoPro}, EPAN achieves the best in SSIM and the second best in PSNR. We would like to point out that Suin \textit{et al.}~\cite{suin2020spatially} achieves the best PSNR by using multi-patch multi-level training, while EPAN reaches comparable performance with the training set of the single-level full image, which is more effective concerning the computational cost.


\section{Conclusion} \label{conclusion}
In this work, we proposed a novel framework for motion deblurring, termed Edge Prior Augmented Networks (EPAN) by leveraging edge information as prior knowledge. EPAN contained not only a content deblurring main branch, but also an edge enhancement auxiliary branch.
Specifically, EPAN employed edge information from the auxiliary branch to augment the deblurring main branch via two components, including an attentive fusion mechanism and an edge-guided loss function. To better verify the effectiveness of motion deblurring algorithms in real scenes, we  curated a dataset, Real Object Motion Blur (ROMB), with paired real sharp and naturally blurry images of fast-moving cars, by using a new dual-camera based image capturing setting. The effectiveness of each component of EPAN was rigorously validated by the ablation study conducted on ROMB. By comparing EPAN with other state-of-the-art approaches on ROMB, RealBlur, and GoPro datasets, we showcased the efficiency of EPAN both qualitatively and quantitatively.

There are several interesting issues that are worth exploring in our future work. Firstly, EPAN is now designed for single images only, but sequence-based data can provide temporal information to further improve motion deblurring, hence we are keen on exploring adding temporal features for EPAN. Secondly, EPAN is developed in a sever-based environment, it will be beneficial to port it to low-cost devices. Thirdly, the ROMB dataset can be extended by adding images with more objects, more light conditions, and more scenarios.


%





\ifCLASSOPTIONcaptionsoff
  \newpage
\fi



\bibliographystyle{IEEEtran}
\bibliography{refs}
\end{document}